\documentclass[11pt,a4paper]{article}
\usepackage[hyperref]{emnlp2020}
\usepackage{times}
\usepackage{latexsym}

\aclfinalcopy % Uncomment this line for the final submission
\usepackage{latexsym}

\usepackage{times}
\usepackage{latexsym} 
\usepackage{multirow}
\usepackage{graphicx}
\usepackage{color, colortbl}
\usepackage{amsmath}
\usepackage{dsfont}
\usepackage{booktabs}       % professional-quality tables

\usepackage{subcaption}

\newcounter{notecounter}
\newcommand{\enotesoff}{\long\gdef\enote##1##2{}}
\newcommand{\enoteson}{\long\gdef\enote##1##2{{
\stepcounter{notecounter}
{\large\bf
\hspace{1cm}\arabic{notecounter} $<<<$ ##1: ##2
$>>>$\hspace{1cm}}}}}
\enoteson
\enotesoff

\def\figref#1{Figure~\ref{fig:#1}}
\def\figlabel#1{\label{fig:#1}\label{p:#1}}

\def\tabref#1{Table~\ref{tab:#1}}
\def\tablabel#1{\label{tab:#1}\label{p:#1}}

\def\secref#1{\S\ref{sec:#1}}
\def\seclabel#1{\label{sec:#1}}
\def\eqref#1{Eq.~\ref{eqn:#1}}

\newcommand{\stackexchange}{SE}
\newcommand{\cdc}{CLF}
\newcommand{\cdknn}{$k-$NN}
\def\bertbase{BERT$_{\scriptsize \textrm{BASE}}$}
\def\bertdapt{BERT$_{\scriptsize \textrm{DAPT}}$}

\usepackage{url}

\definecolor{Gray}{gray}{0.9}
\definecolor{LightCyan}{rgb}{0.88,1,1}

\usepackage{microtype}

\title{Cross-Domain Generalization Through Memorization:\\
A Study of Nearest Neighbors in Neural Duplicate Question Detection}

 \author{Yadollah Yaghoobzadeh*\thanks{*Equal contribution.}, Alexandre Rochette*\footnotemark[1], Timothy J. Hazen\\ {Microsoft Turing} \\\centering{ \texttt{yayaghoo@microsoft.com}}}

\date{}

\begin{document}
%\ninept
%
\maketitle
\begin{abstract}

Duplicate question detection (DQD) is important to increase efficiency of 
community and automatic question answering systems. 
Unfortunately, gathering supervised data in a domain is time-consuming and expensive, and our ability to leverage annotations across domains is minimal. 
In this work, we leverage neural representations and study nearest neighbors for  cross-domain generalization in DQD.  
We first encode question pairs of the source and target domain in a
rich representation space and then using a k-nearest neighbour retrieval-based method, we aggregate the neighbors' labels and distances to rank pairs.
We observe robust performance of this method in different cross-domain scenarios of StackExchange, Spring and Quora datasets, outperforming cross-entropy classification in multiple cases. We will release our codes as part of the publication.
% ervised adaptation to StackExchange domains by self-supervised finetuning of contextualized embedding models like BERt.
%We show the effectiveness of this adaptation in scenarios when source domain comes from different types of distributions.
%Our analysis also reveals that unsupervised domain adaptation on even small amounts of data boosts the performance significantly.
%Further, we show how an approach based on nearest neighbors is effective  for this problem and outperforms training the full model using cross entropy.
\end{abstract}

\section{Introduction}\label{intro}
\begin{figure*}
\centering
\includegraphics[width=1.0\textwidth]{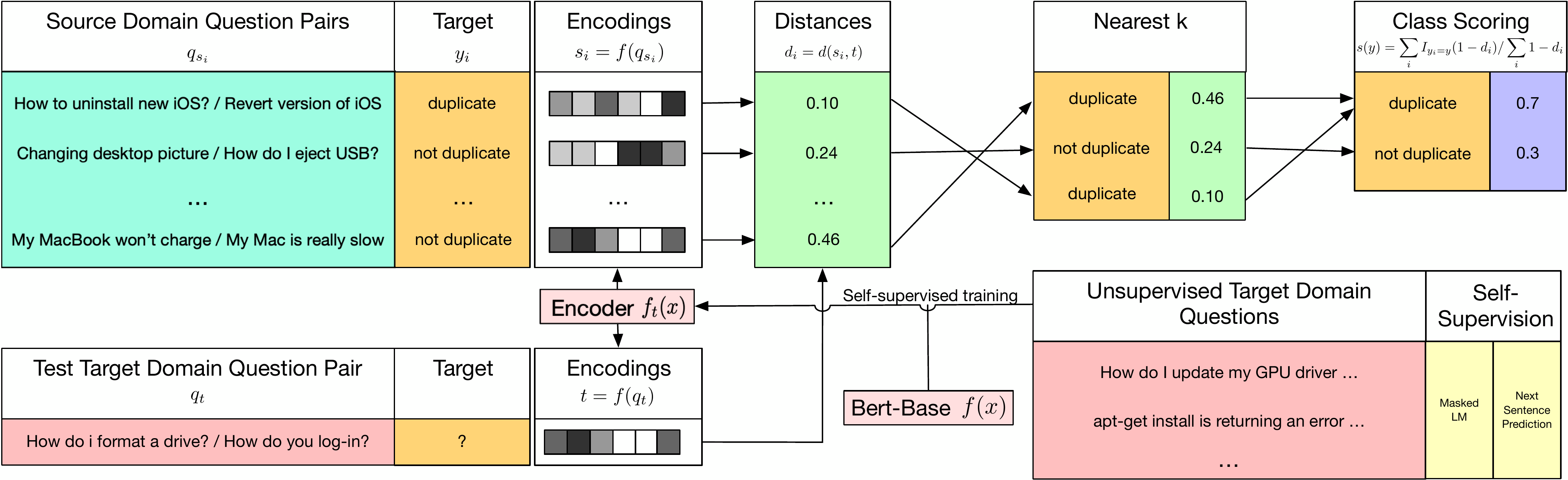}
\caption{An illustration of our \cdknn{}. Question pairs from both Source and Target domains are encoded by in a common representation space, which is 
the result of unsupervised adaptation of BERT on Target data. 
The top k similar items from Source are then aggregated based on their label and distance to provide a final confidence. 
% The  diagram roughly parallels Figure 1 from \citet{khandelwal2019generalization} but adapted to the DQD and cross-domain setup. 
}
\figlabel{knnprocess}
\end{figure*}

Duplicate question detection (DQD) is an important application in information retrieval and NLP \citep{burke1997question,jeon2005finding,lei2016,nakov-semeval2016,ruckle-etal-2019-neural}. It allows systems to recognize when two questions share an answer. This is significant for community forums, such as StackExchange\footnote{https://stackexchange.com/} (\stackexchange{})  to increase their effectiveness in avoiding redundant questions and displaying relevant answers to search questions.
It is also important for FAQ retrieval question answering systems \citep{Sakata19}.

To learn DQD models for \stackexchange{}, question pairs are usually annotated with duplication information that is extracted from community-provided meta-data.
Such annotations are sparse for most domains, e.g., a new \stackexchange{} forum providing support for a new product. 
Therefore, leveraging other training signals either from unsupervised data or
supervised data from other domains is important \citep{shah2018adversarial,poerner-schutze-2019-multi}.

Pre-trained language models (\textbf{PLMs}) like BERT \citep{bert} and RoBERTA \citep{liu2019roberta} are 
great unsupervised textual representations.
Several recent efforts adapt PLMs for the domains of interest  by 
self-supervised fine-tuning on unsupervised domain data, which has shown 
to be promising in several scenarios
 \citep{bioBERT,sciBERT,adaptaBERT2019,gururangan2020don}. 
We follow that and tune BERT on \stackexchange{} domains to obtain richer representations for the task of DQD. 

Recently,  $k$-nearest neighbors ($K-$NNs) is applied on the PLM representations for language modeling \citep{khandelwal2019generalization} and dialogue \citep{fan2020augmenting}.
We extend this line of study and apply \cdknn{} for 
cross-domain generalization in DQD, where the models are trained on data from a \emph{source} domain, and applied on data from a \emph{target} domain. 
To do so, we represent pairs from source and target in a common representation space and then score target pairs using nearest neighbors
in the source pairs. \figref{knnprocess} shows an illustration of this procedure. 

% The specific properties of \stackexchange{} DQD
% is important to make this approach effective.

Our study on AskUbuntu as target and source datasets of \citep{shah2018adversarial}, which include several domains of \stackexchange{} and also Quora and Sprint, reveals that \cdknn{} is more effective compared to cross-entropy classification if (i) the pair representation space from PLMs is rich for the target domain, i.e., adapted on the unsupervised data  from target or similar domains; or (ii)  source and target domains have large distributional shifts. 

We make the following contributions:
(i)  We present the first study of combining strengths of \cdknn{} and 
    neural representations for cross-domain generalization in a sentence matching task, i.e., DQD. 
(ii) Our experimental results  on cross-domain DQD demonstrate that \cdknn{} on 
    rich question-pair representations advances the results of 
    cross-entropy classification, especially when shifts in source to target domains is substantial.

\section{Related Work}
Sparsity in DQD labeled examples in the \stackexchange{} domains is 
 tackled by leveraging the unsupervised data \citep{poerner-schutze-2019-multi}, the supervised data from other domains, or both  \citep{shah2018adversarial,uva2018injecting,ruckle-etal-2019-neural,rochette2019unsupervised}.
We follow these approaches and learn representations from unsupervised data and 
apply them for better generalization when external supervised data in other domains is used.

% Large PLMs \citep{bert,liu2019roberta, lan2019albert,yang2019xlnet} learn high-quality representations from raw text through self-supervised objectives. This ability has been leveraged outside of DQD to provide better cross-domain generalization \citep{bioBERT,sciBERT,adaptaBERT2019}.

A combination of \cdknn{} with neural representations is the subject of several earlier work, mostly in image classification \citep{papernot2018deep,cohen2019dnnKnn} 
\citet{papernot2018deep} show that \cdknn{} is more robust to out-of-distribution examples. 
More related to our work, \citet{khandelwal2019generalization} apply \cdknn{} on neural representations computed from and applied on language modeling task and interpolates its scores with Softmax. They validate that this is effective in different scenarios, including domain adaptation. 
Here we do not compute representations for \cdknn{} on the same task as the one we apply; the representations are computed by language modeling and applied on DQD.

\section{Cross-Domain DQD}
DQD is to identify pairs of questions answered by the same information.  We address DQD for community forums like \stackexchange{}.
In \stackexchange{}, several domains are built to address diverse user needs. We address DQD in a scenario where we perform on a \emph{Target} domain with no task supervised data,  by leveraging its unsupervised text if exists, 
and labeled examples from a \emph{Source} domain.   

% The first of these objectives, masked language modeling, randomly selects a subset of the input tokens to mask and measures its ability to predict them. The second objective is the next sentence prediction, where the ability to predict whether two segments of text follow each other in the original corpus is evaluated.

\subsection{Question Pair Representations}
Our work leverages BERT \citep{bert}, a transformer-based language model, pre-trained on general text from Wikipedia and BookCorpus. 
We fine-tune BERT using its self-supervised objectives on unlabeled questions in \stackexchange{} formatted as (Title, Body). 
Specifically, we concatenate the Title and Body of the questions of a domain to form documents for BERT.
We adopt the terminology of \citet{gururangan2020don} and call this
process \emph{domain-adaptive pre-training (\textbf{DAPT})} and the resulting 
model \textbf{\bertdapt}. 
DAPT tailors the representation towards the particular vocabulary, syntax, and semantics of the problem space (DQD). It also adapts the representation to a specific domain (e.g. AskUbuntu). 
Therefore, we expect \bertdapt{} to produce richer initial representations
for our data.

% The \stackexchange{} question format contains a Title and a Body. 
% The two are related to each other, e.g., Body details the Title, and are shown to contain useful training signals for DQD  \citep{ruckle-etal-2019-neural}.
% We use 
% unlabeled questions of Target in (Title, Body) format to produce rich \bertdapt{} representations.
% We hypothesize that \bertdapt{} representations are  rich for DQD task \enote{yy}{TODO: why?}

% Also notably since the first sentence of body usually 
% explains the title in other forms, this could be seen 
% as some training signal, as it is used in \citet{ruckle-etal-2019-neural}.
 
To obtain a representation for a pair of questions $(q_1, q_2)$,
we concatenate the text of both items and feed that as input to \bertdapt. The two items are separated by the [SEP] token. We regard the embedding of the first token ([CLS]) at the final layer as our pair representation.  
This representation is then 
utilized in two ways to produce DQD predictions, as described in the following. 
% The first use is as an initialization for a supervised classification model trained on the $S$ domain data. The second is to use the representation directly to identify the k-nearest neighbors from the $S$ data and infer its label from them. 

\subsection{Classification (\cdc{})}
For \cdc{}, we follow the standard BERT training for sequence classification starting from \bertdapt{} as a better initial PLM
for cross-domain DQD.
In this setting, the CLS embedding is the input to a classification layer with a cross-entropy loss.
The gradients are back-propagated to the \bertdapt{} parameters through CLS embedding and tuned for the DQD task.

\subsection{$k-$Nearest Neighbors (\cdknn{})}
We leverage the self-supervised representation $f_{t}$, 
corresponding to the CLS embedding in \bertdapt{}  and encode each pair $q_{s_i}$ in the Source training set $D_s$ using $f_t$ and preserve its associated label $y_i \in  \{duplicate, \neg duplicate\}$, as illustrated in \figref{knnprocess}.
A distance function $d$ between two vectors $u$ and $v$ is selected in order to establish the nearest neighbors. We use the cosine distance for this purpose: $d(u, v) = 1 - cosine(u, v)$ %\frac{ u \cdot v } {\norm{u}_2\norm{v}_2}

One score for each potential label $y$ of a test pair $q_{t}$ is then computed using $s$, representing the fraction of the mass of $1-d_i$ in K of each.
\begin{equation}
    s(y) = \frac{ \sum_i^{k}{\mathds{1}_{y_i=y}(1-d(f_t(q_{s_i}), f_t(q_{t}))}} {\sum_i^{k}{1-d(f_t(q_{s_i}), f_t(q_{t}))}}
\end{equation}

% The final score for class $y$ can be thresholded to obtain a classification decision.

\section{Datasets}
\begin{table}[h]
% \footnotesize
\resizebox{\columnwidth}{!}
{\begin{tabular}{lllll}
\toprule
\textbf{Dataset}    & \textbf{Questions} & \textbf{Train} & \textbf{Dev}   & \textbf{Test}  \\ \hline
AskUbuntu  & 305,769      & 9,106 & 1,000 & 1,000 \\
SuperUser  & 390,378      & 9,106 & - & - \\
Sprint     & 31,768 & 9,106 & - & -  \\
Quora      & 537,211 & 9,100 & - & -\\
\bottomrule
\end{tabular}}
\caption{ 
The number of duplicates taken from \citet{shah2018adversarial}.
For AskUbuntu, as our Target, Train, Dev and Test, 
and for others Train numbers are shown. 
Note that all AskUbuntu questions are considered for \bertdapt{}.
}
\label{tab:datasets}
\end{table}

We experiment on cross-domain DQD datasets of \citet{shah2018adversarial}\footnote{\url{github.com/darsh10/qra_code}} (See \tabref{datasets}). 
% It includes AskUbuntu and SuperUser from \stackexchange{}
% as well as Quora and Sprint.
For AskUbuntu and SuperUser, the positive examples are taken from the duplicate marks in \stackexchange{}. For Sprint, three paraphrases are generated by annotators for each question in a set of FAQ.
In these three datasets, 100 negatives are sampled randomly per each positive. 
The annotation of Quora comes from the released
Quora question pairs dataset \citep{qqp}.

We further extract all questions for \stackexchange{} domains 
from the dump files.\footnote{archive.org/details/stackexchange}.
These are integrated in our unsupervised adaptations.
For our analysis in \secref{analysis}, we create two additional unsupervised corpora from \stackexchange{}.
The first is from Academia consisting of  around 27K questions. The second is from 33 different \stackexchange{} domains (See \tabref{33domains} in Appendix), composed of around 1.5M questions.

\subsection{Lexical Similarity Statistics}
\label{sec:jaccard}

\begin{table}
\centering
\begin{tabular}{lcc|c}
\toprule
& \textbf{dup}     & \textbf{$\neg$dup}        & $S$(AskUbuntu)\\
\midrule
AskUbuntu & 0.16        & 0.03        & 1.00     \\
SuperUser & 0.19        & 0.03        & 0.22 \\
Sprint     & 0.37	& 0.04 & 0.03\\
Quora     & {0. 47}        & {0.30} & 0.12 \\
\bottomrule
\end{tabular}
\caption{Lexical similarity (measured by Jaccard index) between question pairs within domains grouped by classes (first two columns) and between vocabulary of Sources and AskUbuntu.  
}
\tablabel{domain_distances}
\end{table}
We select AskUbuntu as our only Target.
In \tabref{domain_distances} (last column), we show the lexical similarity between 
each Source and the Target.     
Accordingly, Sprint and Quora hold low similarity with AskUbuntu, and SuperUser is the most similar domain.

We also present the similarity between paired questions in each class of duplicate and non-duplicate in \tabref{domain_distances}.
We observe that a duplicate pair in Sprint and Quora has higher word-overlap on average compared to \stackexchange{} datasets. 
Quora has another significant difference: its negative pairs are selected to have a high lexical overlap. 
% , while for other datasets the negatives are chosen randomly.
This means that the labeling function in Quora is different from others.

\section{Experiments}
To obtain \bertdapt, we fine-tune \texttt{bert-base-cased}
(\bertbase) using language modeling scripts in 
Transformers \citep{Wolf2019HuggingFacesTS} for $3$ epochs using default hyperparameters.
To clarify the effects of DAPT, we experiment with \bertbase{} as well.
For \cdc{}, we fine-tune BERT on task training data for $10$ epochs with a learning rate of 5e-5, early stopping on the Target dev set.
All our \cdknn{} experiments are done with  with the $k=100$ using Faiss library \cite{JDH17}.

% The comparable result in \citet{ruckle-etal-2019-neural} from a BiLSTM model trained on a large set of (Title, Body) pairs is also added.

\paragraph{Evaluation metric}
Since the annotations are incomplete in \stackexchange{}, \citet{shah2018adversarial} propose to use AUC as the \textbf{metric} for DQD performance.
They report the normalized AUC(.05), which is the area under the curve of the true positive rate as function of the false positive rate ($fpr$), from $fpr = 0$ to $fpr = .05$. We follow the same protocol and use AUC(.05) metric. 

\subsection{Results}

\begin{table}[h]
\setlength{\tabcolsep}{1.5pt}
\small
\centering
\begin{tabular}{p{0.005cm} lcccc}
\toprule
& & \multicolumn{4}{c}{Target = AskUbuntu} \\
& \textbf{Model}            
& \textbf{AskUbuntu} & \textbf{SuperUser} & \textbf{Sprint} & \textbf{Quora}  \\
\midrule
& \multicolumn{5}{l}{\emph{\bertdapt}}\\
1 &  \hspace{0.1cm} \cdc{}    
& .923                         & .870 
& .749                      & .609 \\
2 & \hspace{0.1cm} \cdknn{}             
& \textbf{.936}                         & \textbf{.908}
& \textbf{.753}                      & \textbf{.800}                      \\
\midrule
 & \multicolumn{5}{l}{\emph{\bertbase}}                                         \\
3 & \hspace{0.1cm} \cdc{}
& .899                 & .779 
& .562                      & .515                      \\
4 & \hspace{0.1cm} \cdknn{}     
& .871                 &  .755
& .649                      & .621                      \\

\midrule \midrule
 &  \multicolumn{4}{l}{\emph{From \citet{shah2018adversarial}}}\\
5 & \hspace{0.1cm} BiLSTM 
& .858 & .796 
& .615 & .446 \\ 
% \midrule
% &  \multicolumn{4}{l}{\emph{From \citet{ruckle-etal-2019-neural}}} \\
% 6 & \hspace{0.1cm} WS-TB & .871 & - & -  \\
\bottomrule
\end{tabular}
\caption{Comparing AUC(.05) results of our models and the baseline.
Four Sources are evaluated for the Target, i.e., AskUbuntu.
\bertdapt{} is the adaptation of \bertbase{} on AskUbuntu unsupervised data.}
\tablabel{mainresults}
\end{table}

In \tabref{mainresults}, we present the performance of our models for AskUbuntu evaluation set
given Source data from AskUbuntu, SuperUser, Sprint, or Quora.
We add in-domain (AskUbuntu as Source) results for comparison.
We include the results of \citet{shah2018adversarial} obtained on the same data by learning domain-adversarial BiLSTM models in line 5. 

The first block corresponds to \bertdapt{}: the adapted BERT on AskUbuntu unsupervised data. We see that \cdknn{} outperforms \cdc{} in all cases (line 2 vs. 1), confirming that \cdknn{}
is more robust if the pair representation is rich.
The most obvious improvement belongs to Quora as Source (.609 to .800), where the labeling function shifts significantly (See \secref{jaccard}).

In the second block, \bertbase{} results in consistently worse models (line 3-4)
compared to  \bertdapt{}.
Here for Sprint and Quora, \cdknn{} again outperforms \cdc{},
giving more evidence about robustness of \cdknn{} in the case of 
domain shifts. 
However, for SuperUser, a closely related domain
to AskUbuntu and also AskUbuntu itself, 
\cdknn{} underperforms.
% labeling function
% is also pretty similar (cf., \secref{jaccard}).
Given that the input representations to \cdknn{} are not tuned on \stackexchange{} in the case of \bertbase{},
this is not surprising. \cdc{} fine-tunes the representations as part of its task training on the Source data, which in this case is a related or same domain as the Target.

% \begin{table}[h]
% \centering
% \begin{tabular}{p{0.005cm} lcc}
% \toprule
% & \textbf{Unsupervised Data} &  \textbf{\cdc} &  \textbf{\cdknn{}}    \\
% \midrule
% 1 & None                &  .839       &   .813      \\
% 2 & Source              & .889        &   .911     \\ % 
% 3 & Target              & .897        &   .922     \\ % 
% 4 & Other (Academia)    & .839        &   .836 \\
% 5 & 33 \stackexchange{} domains  & \textbf{.917}        &   \textbf{.930}   	\\ 
% \bottomrule
% \end{tabular}
% \caption{Results on AskUbuntu as Target. 
% Average AUC(.05) over Source $\in$ \{AskUbuntu, SuperUser\} for \cdc{} and \cdknn{} as a function of the unsupervised dataset used
% for \bertdapt. None corresponds to \bertbase.}
% \tablabel{results_lm_variation}
% \end{table}

\begin{table}[h]
\centering
\begin{tabular}{p{0.002cm} lcc}
\toprule
& \textbf{Unsupervised Data} &  \textbf{\cdc} &  \textbf{\cdknn{}}    \\
\midrule
1 & None                & .779       &   .755      \\
2 & Source              & .855        &   .886     \\ % 
3 & Target              & .870        &   .908     \\ %
4 & Unrelated           & .778        &   .782 \\
5 & 33 \stackexchange{} domains  & \textbf{.891}        &  
\textbf{.917}   	\\ 
% \midrule
% 6 & Target (16MB)              & -        &   .811     \\ % 
% 7 & Target-OnlyTitle (16MB)              & .767       &   .773     \\ % 
\bottomrule
\end{tabular}
\caption{AUC(.05) of training on Source = SuperUser and
evaluating on Target = AskUbuntu, as a function of the input corpus
for \bertdapt. None corresponds to \bertbase{} and Unrelated to Academia.
(Similar results but with Source = AskUbuntu is in Appendix, \tabref{results_lm_variation_ask}.)}
\tablabel{results_lm_variation}
\end{table}

\subsection{Domain of Unsupervised Data in \bertdapt{}}
\seclabel{analysis}
Here we aim to understand more about the impact of the domain of unsupervised data on the quality of \bertdapt{}.
In \tabref{results_lm_variation}, we report results for SuperUser as Source and AskUbuntu as Target, and vary the unsupervised corpus, starting from no data (i.e., \bertbase{}).

We choose these domains (lines 2-5): Source (SuperUser), Target (AskUbuntu), Unrelated (Academia) as a lexically distant domain to Target, and a set of 33 \stackexchange{} domains including Source and Target (See \tabref{33domains} for full list).

\tabref{results_lm_variation} demostrates that adaptation on Target data (line 3) is better
than either of Source (line 2) or the unrelated domain (line 4).
Adaptation on a large number of domains (line 5) is the best; the information across a diverse set of domains is complementary for the task.
Notably, we observe that the representation is more critical for \cdknn{} compared to \cdc{}: the difference between the best representation (33 domains) and worst (\bertbase) is much greater in \cdknn{} compared to \cdc{}. This behavior is understandable as \cdc{} further updates the representation during its task training, while for \cdknn{}, it remains fixed.

\section{Conclusion}
In this work,  we studied applying \cdknn{} in DQD cross-domain generalization.
We compared \cdknn{} and a cross-entropy classifier when different 
question-pair representations are available.
Our results showed that domain-adaptive pre-training on target data gives rich representations, and \cdknn{} is more robust against distributional shifts compared to classification if question pairs are encoded by these rich representations.

We plan to extend our study to other tasks and understand better  the strengths of memorization in learning robust models where rich PLM embeddings are utilized to represent examples. 
We believe concurrently that the promising results and findings of this presented study could benefit other NLP research to explore this direction more.

\bibliography{refs}
\bibliographystyle{acl_natbib}

\appendix

\begin{table*}[h]
\centering
\small
\begin{tabular}{p{2.5in}l}
     academia
android
anime
apple
askubuntu
astronomy
aviation
bicycles
biology
bitcoin
boardgames
buddhism
chemistry
christianity
cogsci
cooking
crypto
cs
gaming
hinduism
islam
judaism
linguistics
mechanics
meta.stackexchange
meta.superuser
philosophy
politics
skeptics
sports
superuser
unix
workplace
 
\end{tabular}
    \caption{The 33 StackExchange domains used in our unsupervised BERT adaptation analysis.}
    \label{tab:33domains}
\end{table*}

\begin{table}[h]
\centering
\begin{tabular}{p{0.002cm} lcc}
\toprule
& \textbf{unsupervised Data} &  \textbf{\cdc} &  \textbf{\cdknn{}}    \\
\midrule
1 & None                & .899       &   .871      \\
3 & Target              & .923        &   .936     \\ %
4 & Unrelated           & .899        &   .890 \\
5 & 33 \stackexchange{} domains  & \textbf{.942}        &  
\textbf{.942}   	\\ 
\bottomrule
\end{tabular}
\caption{AUC(.05) of training on Source = AskUbuntu and
evaluating on Target = AskUbuntu, as a function of the input corpus
for \bertdapt. None corresponds to \bertbase{} and Unrelated to Academia.}
\tablabel{results_lm_variation_ask}
\end{table}

% \input{others/bigtable}

% \begin{table*}[]
% \centering
% \begin{tabular}{lllrcc}
% \toprule
%   & $d_1$   & $d_2$   & \multicolumn{1}{c}{Example}                     & \bertbase & \bertdapt \\
%   \midrule
% 1 &  low  & low  & SuperUser $\rightarrow$ AskUbuntu  & CLF   & KNN              \\
% 2 & low  & high & AskUbuntu* $\rightarrow$ AskUbuntu &  ?  & ?              \\
% 3 & high & low  & Sprint $\rightarrow$ AskUbuntu   & KNN   & KNN            \\
% 4 & high & high & Quora $\rightarrow$ AskUbuntu      & KNN   & KNN            \\
% \bottomrule
% \end{tabular}
% \caption{$s_1$: lexical distance . $d_2$: distance in labeling function for negatives. }
% \end{table*}

% \begin{figure*}[ht]
% \centering
% \includegraphics[width=1\textwidth]{figures/knn-figure-version3.eps}
% \caption{KNN Test Graphic here}
% \label{fig:knnprocess_app}
% \end{figure*}
% \begin{figure*}[ht]
% \centering
% \includegraphics[width=1\textwidth]{figures/uda.eps}
% \caption{UDA Test Graphic here}
% \label{fig:udaprocess_app}
% \end{figure*}

\end{document}